%% file: arXiv.tex
\documentclass{article} 
\usepackage[a4paper,headheight=20pt,top=25mm,bottom=35mm,left=35mm,right=25mm]{geometry}
\usepackage{arxiv}
\usepackage[T1]{fontenc}    
\usepackage{hyperref}       
\usepackage{amsfonts}       
\usepackage{amsmath}
\usepackage{fancyhdr}       
\usepackage{graphicx}       
\usepackage{amsthm}
\usepackage{hyphenat}
\usepackage{url}
\usepackage{subfigure}
\usepackage{enumitem}
\newtheorem{assumption}{Assumption}
\graphicspath{{img/}}

\title{Can Interpretable Reinforcement Learning Manage Prosperity Your Way?}

\author{Charl Maree%
\thanks{Second affiliation: Chief Technology Office, Sparebank 1 SR-Bank, Stavanger, Norway.} 
\, and Christian Omlin \\ 
Center for Artificial Intelligence Research\\ 
University of Agder\\ 
Grimstad, Norway \\ 
\texttt{\{charl.maree,christian.omlin\}@uia.no} \\ }

\begin{document}
\maketitle
\input{source/abstract}
\input{source/introduction}
\input{source/related_work}
\input{source/methodology}
\input{source/results}
\input{source/conclusions}
\input{source/statements}

\bibliographystyle{unsrt}
\bibliography{references}
\end{document}

%% file: source/abstract.tex
\newcommand{\myabstract}{
Personalisation of products and services is fast becoming the driver of success in banking and commerce. Machine learning holds the promise of gaining a deeper understanding of and tailoring to customers' needs and preferences. Whereas traditional solutions to financial decision problems frequently rely on model assumptions, reinforcement learning is able to exploit large amounts of data to improve customer modelling and decision-making in complex financial environments with fewer assumptions. Model explainability and interpretability present challenges from a regulatory perspective which demands transparency for acceptance; they also offer the opportunity for improved insight into and understanding of customers. Post-hoc approaches are typically used for explaining pretrained reinforcement learning models. Based on our previous modeling of customer spending behaviour, we adapt our recent reinforcement learning algorithm that intrinsically characterizes desirable behaviours and we transition to the problem of asset management. We train inherently interpretable reinforcement learning agents to give investment advice that is aligned with prototype financial personality traits which are combined to make a final recommendation. We observe that the trained agents' advice adheres to their intended characteristics, they learn the value of compound growth, and, without any explicit reference, the notion of risk as well as improved policy convergence.
}

 \begin{abstract}
     \myabstract{}
 \end{abstract}


\keywords{AI in banking; personalized services; asset management; explainable AI; reinforcement learning; policy regularisation}

%% file: source/introduction.tex
\section{Introduction}
Financial service providers are employing ever-advancing methods to improve the level of personalisation of their services \cite{stefanel19, Jaiwant22}. Artificial intelligence (AI) is a promising tool in this pursuit in areas such as anti-money laundering, trading and investment, and customer relationship management \cite{vdBurgt19}. Examples of \emph{personalised} services are recommender systems for product sales \cite{oyebode20}, risk evaluation for credit scoring \cite{Bhatore20}, and segmentation for customer-centric marketing \cite{Desai22}. More commonly, AI has been applied to stock trading via ensemble learning \cite{Jothimani19}, currency recognition using deep learning \cite{Zhang19}, stock index performance through time-series modelling with feature engineering \cite{hsu21}, and investment portfolio management using reinforcement learning (RL) \cite{Kolm19, Fischer2018}. These applications generally lack the personalisation needed to enhance customer relations and support service delivery for growing customer bases. The lack of explainability and interpretability has thus far hindered the wider adoption of machine learning, mainly due to model opacity; model understanding is essential in financial services \cite{arrieta2020, Cao21, maree_towards}. We distinguish between explainability and interpretability: explainability refers to a symbolic representation of the knowledge a model has learned, while interpretability is necessary for reasoning about a model's predictions. 

We have previously investigated the \emph{interpretability} of systems of multiple RL agents \cite{maree_yourway}. A regularisation term in the objective function imposed a desired agent behaviour \emph{during} training. For our current purpose of asset management, our agents learn distinct optimal policies for continuously distributing a fixed monthly amount across five assets: a savings account, property, stocks, luxury items, and mortgage repayments. It is our intention to align the agents' characteristics and behaviour with personality traits - openness, conscientiousness, extraversion, agreeableness, and neuroticism - as proposed for modelling spending behaviour in \cite{Gladstone2019}. A linear combination of the resulting policies can provide investment advice that matches each customer's unique personality profile. This intrinsic interpretability may fulfill the promise of a digital private assistant for personal wealth management.

%% file: source/related_work.tex
\section{Related Work}
Recent evidence has revealed a causal relationship between spending patterns and individual happiness \cite{Matz2016}: we are happiest when our spending matches our personality. For instance, extraverted individuals typically prefer spending at a bar rather than at a bookshop, while the opposite may apply to introverts. Our premise is that spending personality traits can be carried over to asset management: we are happiest when our investment matches our personality. For instance, conscientious investors may prefer the predictability of property over the volatility of stocks. This is consistent with the high affinity of conscientious spenders towards residential mortgages \cite{Matz2016}. It is compelling to expand the notion of personality traits from spending to wealth creation, i.e., to base personal investment advice on historical spending behaviour \cite{maree_clustering, maree_understanding}.

RL has been extensively applied to stock portfolio management \cite{Bartram21, Jurczenko20, Lim21, Pinelis22, millea2021, maree_balancing}, but not yet to \emph{holistic} asset management; the lack of model transparency may be a contributing factor. Interpretation of RL agents typically follows model training \cite{heuillet2021a, wells21, gupta21}; our ambition is to \emph{impose} a desired characteristic behaviour \emph{during training}, thus making it an intrinsic property of the agent. Based on a prior that defines a desired behaviour, we extend the deep deterministic policy gradient (DDPG \cite{lillicrap2019}) objective function with a regularisation term \cite{maree_yourway}. Formally, for each agent $i$, this objective function is given by:
\begin{align} \label{eqn:regularized_obj_func}
    &J(\theta_i) = \mathbb{E}_{o_i,a_i \sim \mathcal{D}} \left[ R_i(o_i,a_i) \right] - \lambda L_i \\
    &L_i = \frac{1}{M_i} \sum_{j=0}^{M_i} \left[ \mathbb{E}_{a \sim \pi_{\theta_i}}(a_j) - (a_{j} \vert \pi_{0_i}(a)) \right]^2 \nonumber
\end{align}
where $\theta_i$ is a set of parameters governing the policy, $\mathcal{D}$ is the replay buffer, $R_i(o_i,a_i)$ is the reward for action $a_i$ with the partial sate observation $o_i$, $\lambda \in \mathbb{R}_{\geq 0}$ is a scaling parameter, $M_i$ is the number of actions, and $\pi_{0_i}$ is the prior that defines the desired behaviour of the agent. Note that the prior is independent of the state, which simplifies it and thus makes it interpretable; this is a departure from traditional policy regularisation methods such as KL-regularisation and entropy regularisation which aim to improve learning convergence instead \cite{ziebart2010, haarnoja2017}. Traditional regularisation encourages state space exploration by increasing the entropy of the policy, whereas our method guides agents' learning towards the prior and thus imposes a desired characteristic behaviour.

%% file: source/methodology.tex
\section{Empirical Methodology}
The aim of this work was to create an interpretable AI for personal investment management. We selected five assets in which a customer could invest a monthly amount over a duration of 30 years: a savings account, property, a portfolio of stocks, luxury expenditures, and additional mortgage payments. We include luxury expenditure to the portfolio under the premise that it may increase customer satisfaction in their portfolios \cite{Matz2016}. We define luxury items as any expenditure that may appeal to a person's personality profile; people scoring high on openness might derive joy from spending money on travelling, people scoring high on extraversion may prefer to spend money on festivities with other people \cite{Matz2016}, while other luxury items such as cars or artwork are also possible. We modelled the growth rates of assets according to historical index data, which we describe below.

\subsection{Modelling Assumptions}
\begin{figure}[!hb]
    \centering
    \includegraphics[width=0.7\linewidth]{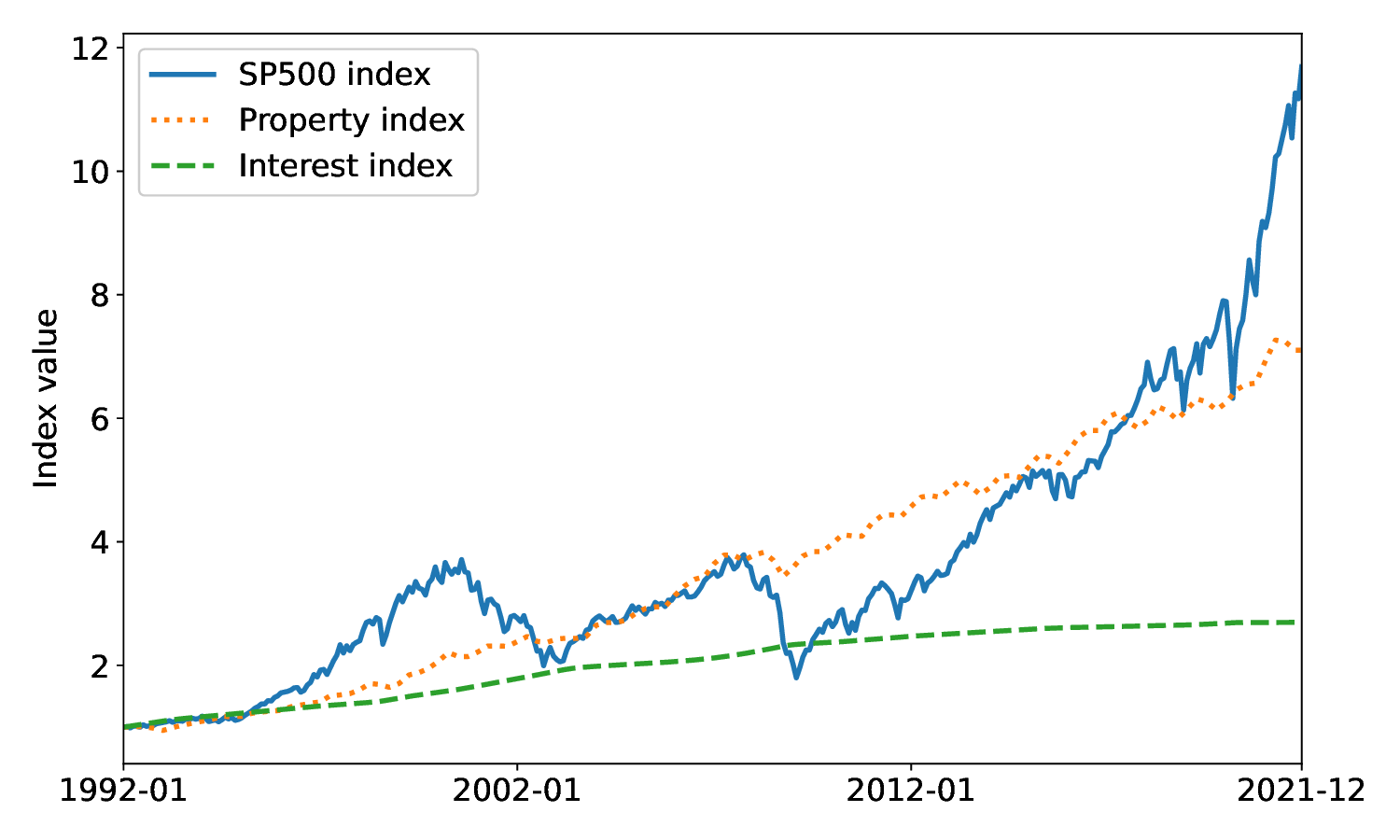}
    \caption{Three asset value indices for a period of 30 years: The S\&P~500 stock index, the Norwegian property index, and the Norwegian interest rate index. All indices are relative to their respective values on 1 January 1992. While the stock index performs the best overall, it has the highest volatility and therefore the highest risk. Conversely, the interest rate index has the lowest risk but also the lowest growth.}
    \label{fig:data}
\end{figure}

We continuously distribute funds into assets based on the indices of the S\&P~500 \cite{yf_sp500}, Norwegian property \cite{ssb_prop}, and the Norwegian interest rate \cite{nb_int}. In addition, we invest in mortgages and luxury items. We show this data for a 30-year period in Figure~\ref{fig:data}.

We make a number of assumptions which limit the scope of the portfolio and simplify investment choices to make the characterization of agent behaviour and interpretation of investment strategies tractable.

\begin{assumption} \label{ass:growth}
  Asset growth rates can be modelled by their respective asset indices, i.e., a stock portfolio may be modeled by a major stock index - e.g., the S\&P~500 -, and an investment in property by its corresponding index. 
\end{assumption}
The outright investment in indices such as S\&P~500 is very common; it will return the growth rates according to these indices. This is a conservative assumption as stock portfolio optimization frequently outperforms indices, which may serve as a performance measure of the investment strategy \cite{maree_balancing}.

To give personalised advice, we depart from the premise that there is a mere correlation between spending behaviour and happiness. We are expanding the notion of the causal relationship of spending patterns and customer satisfaction to chart an investment strategy and provide advice that is aligned with customer personality \cite{Matz2016}. 
We rated assets in terms of their risk and expected return from the historical values of the indices, and their liquidity, capital requirements, and novelty from domain experts. They also scored the different personality types' affinity for different assets from the interval $[-1,+1]$. Individuals with a high degree of openness may prefer the liquidity of stocks over mortgage payments, whereas neurotic individuals may prefer savings accounts over stocks in their portfolio. These coefficients, shown in Table~\ref{tab:coefficients}, are the weighted sums of the asset ratings and affinities.

\begin{table}[!ht]
 \centering
    \caption{Coefficients relating asset risk, expected return, liquidity, capital requirement, and novelty to prototypical personality traits: openness (O), conscientiousness (C), extraversion (E), agreeableness (A), and neuroticism (N). The values are in the range $[-1,1]$.} \label{tab:coefficients}
    \footnotesize
    \begin{tabular}{l|ccccc}
        Investment & O & C & E & A & N \\ \hline
        Savings  & -0.11 &  0.08 & -0.15 &  0.51 &  0.68 \\
        Property & -0.15 &  0.32 & -0.22 & -0.36 & -0.24 \\
        Stocks   &  0.82 & -0.61 &  0.95 &  0.42 &  0.12 \\
        Luxury   &  0.16 & -0.51 & -0.07 & -0.80 & -0.81 \\
        Mortgage & -0.72 &  0.72 & -0.52 &  0.23 &  0.25 
    \end{tabular}
\end{table}

We define a Markov decision process (MDP) for a multi-agent RL setting. The \emph{states} consist of customer age\footnote{Customer age is normalized to a range of $[0,1]$}, the values of the assets\footnote{Asset values are scaled by $1:10^6$}, and two market indicators for each of the three indices, i.e., their mean asset convergence divergence (MACD)\footnote{MACD here is the difference between the 26-month and the 12-month exponential moving average of a trend.} which predicts trend reversals and relative strength index (RSI)\footnote{$RSI = 100 - 100 / (1+\frac{P_x}{N_x})$ where $P_x$ and $N_x$ are the average positive and negative changes to the index values respectively, for $x$ periods.} which corrects for potential false predictions by MACD. \emph{Rewards} are the changes in portfolio values between time steps and \emph{actions} are the continuous distribution of funds across the portfolio of assets. We make an initial loan of 2 million NOK in a mortgage and monthly investments over 30 years totalling 3.34 million NOK:

\begin{assumption} \label{ass:initial}
  The initial values for a portfolio consist of a mortgage of NOK 2 million and a property valued at NOK 2 million. All other assets have zero initial value.
\end{assumption}
It is easy to adjust these initial portfolio assignments for different individuals.

\begin{assumption} \label{ass:monthly_investment}
  We make consistent monthly investments of 10~000 Norwegian kroner (NOK).
\end{assumption}
This can be easily modified for individual customers' contributions.

There is a priori no lower limit on the investment amounts: 

\begin{assumption} \label{ass:real_estate}
  Property investment does not require bulk payments, i.e., smaller investments can be made through property funds, trusts, or crowdfunding.
\end{assumption}
While investment in physical real estate normally requires larger deposits, we allow our agents to invest smaller amounts into the property market, i.e., a fraction of the monthly investment contribution specified in Assumption~\ref{ass:monthly_investment}. This is not a strong assumption as it is possible to invest smaller amounts in property indices, trusts, funds, etc.

We assign interest rates for savings accounts at 5-10\% below, and those of mortgage accounts at 5-10\% over the interest index. Individuals younger than 35 years receive the more beneficial interest rate, as is common in Norwegian banks. Luxury items experience a depreciation of 20\% per year; the depreciation of luxury items is highly variable and depends on the item, e.g., while artwork may appreciate, cars typically depreciate rapidly:
\begin{assumption} \label{ass:lux_depr}
  Luxury items depreciate at 20\% per year.
\end{assumption}

Dividends are normally included in the calculation of indices and monthly transactions are relatively infrequent compared to high frequency trading:
\begin{assumption} \label{ass:additional_income}
  Any additional income from investments - such as dividend payouts or rental income - as well as costs such as transaction costs and fund management costs are ignored.
\end{assumption}

\subsection{Agents}
We train five DDPG agents, one for each of the five personality traits. Using Equation~(\ref{eqn:regularized_obj_func}) we regularise their objective functions with a prior derived from their respective personality traits in Table~\ref{tab:coefficients}, e.g., the openness prior $\pi_0^O$ places the most weight on stocks and avoids mortgage repayments, property investment, and savings, while the conscientiousness prior $\pi_0^C$ places the most weight on mortgage repayments and avoids stocks and luxury expenditure. These priors, shown in Table~\ref{tab:priors}, are probability distributions across the investment channels and therefore add up to one.

\begin{table}[!ht]
 \centering
    \caption{Regularisation priors $\pi_0^a$ for each agent $a \in $ \{openness (O), conscientiousness (C), extraversion (E), agreeableness (A), and neuroticism (N)\}.} \label{tab:priors}
    \footnotesize
    \begin{tabular}{l|ccccc}
        Investment & $\pi_0^O$ & $\pi_0^C$ & $\pi_0^E$ & $\pi_0^A$ & $\pi_0^N$ \\ 
        \hline
        Savings  & 0.00 & 0.07 & 0.00 & 0.44 & 0.64 \\
        Property & 0.00 & 0.28 & 0.00 & 0.00 & 0.00 \\
        Stocks   & 0.84 & 0.00 & 1.00 & 0.36 & 0.12 \\
        Luxury   & 0.16 & 0.00 & 0.00 & 0.00 & 0.00 \\
        Mortgage & 0.00 & 0.65 & 0.00 & 0.02 & 0.24 
    \end{tabular}
\end{table}

Agents' actor and critic neural networks each consist of two fully connected feed-forward layers with 2000 nodes in each layer. The actor networks each have a final soft-max activation layer while the critic networks have no final activations. We tuned the hyperparameters using a one-at-a-time parameter sweep resulting in learning rates of $0.004$ and $0.001$ for the actors and critics respectively, target network update parameters of $\tau=0.05$, and regularisation coefficients of $\lambda=2$. Training batch sizes were 256 time steps and we sized the replay buffer to hold 2048 transitions. Each iteration collected 256 time steps and completed two training batches.

%% file: source/results.tex
\section{Results}
Each of our investment agents learns an optimal investment strategy for their respective prototypical personality traits, for instance, openness. The final portfolio values after 334 months of investing according to these policies are shown in Table~\ref{tab:portfolio_val}. Given the common total investment of 3.34 million NOK, the compound annual growth rate varies between 5.8\% and 7.8\% which is the maximum return possible if investing in stocks only.

\begin{table}[ht]
    \centering
    \footnotesize
    \caption{Portfolio values of the five optimal policies for each of the prototypical personality traits.} \label{tab:portfolio_val}
    \begin{tabular}{l|c}
        Policy & Final portfolio value (NOK 1M) \\
        \hline
        Openness & 22.4 \\
        Conscientiousness & 18.8 \\
        Extraversion* & 27.7 \\
        Agreeableness & 20.5 \\
        Neuroticism & 16.4 \\
        \hline \hline
        Personal agent & 20.3 \\
    \end{tabular} \\
    \vspace{3pt}
    \begin{minipage}{.9\linewidth}
        \footnotesize{*This agent's regularisation prior was coincidentally the same as the optimal monetary policy $\pi^M$} and it achieved the maximum possible final portfolio value.
  \end{minipage}
\end{table}

\begin{figure}[!ht]
    \centering
    \includegraphics[width=0.7\linewidth]{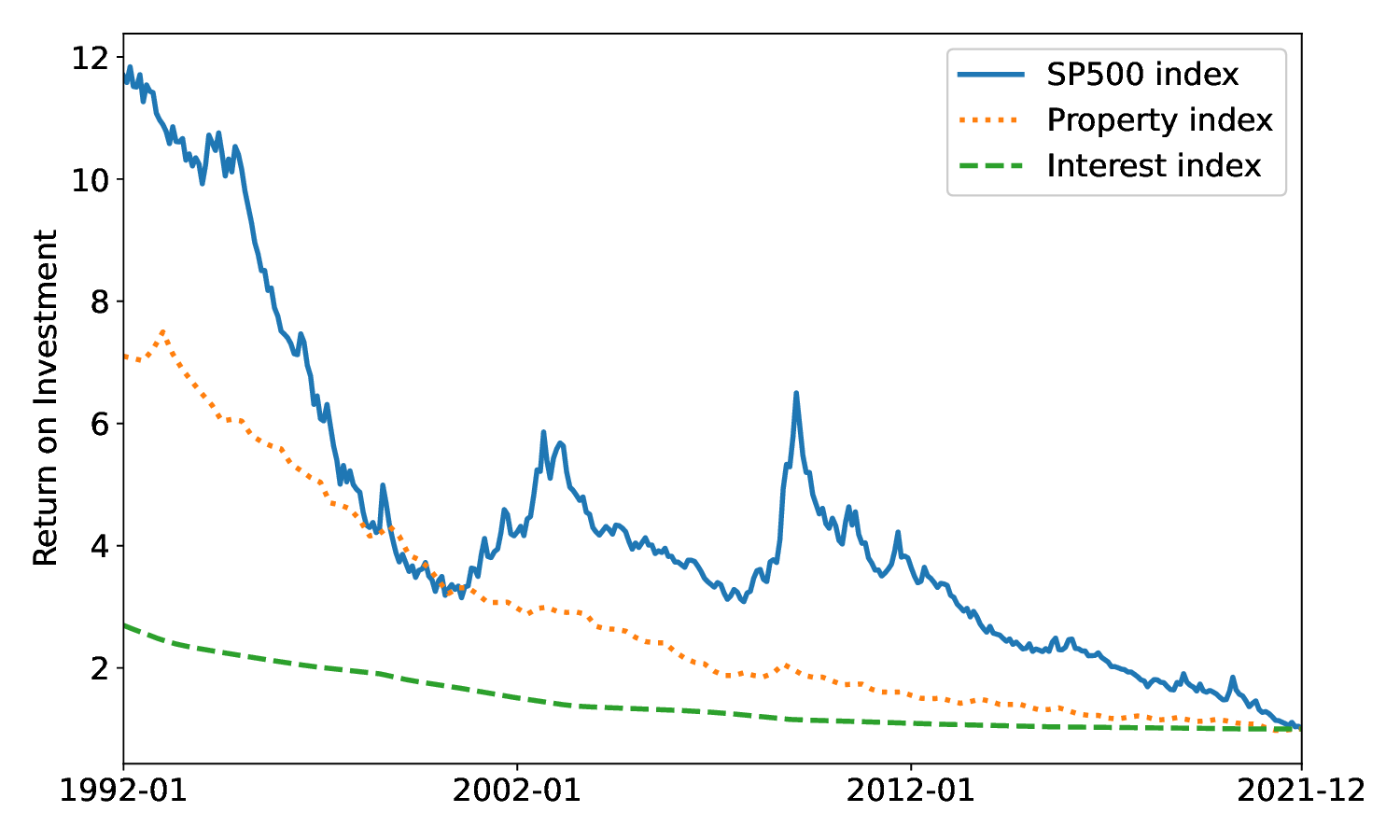}
    \caption{The return on investment at every time step, calculated as the index value at the final time step divided by the index value at the current time step. It is clear that S\&P~500 has the greatest return on investment at every time step, except for a brief period in ca. 2000 where it was marginally below the property index. Therefore, the optimum monetary policy $\pi^M$ is to always invest the maximum amount into stocks.}
    \label{fig:optimum_policy}
\end{figure}

Note that these personalised policies did not achieve the same final portfolio value. In fact, the optimum policy in monetary terms $\pi^M$ in this case would have been to always buy stocks as shown in Figure~\ref{fig:optimum_policy}; this is the default policy an agent will converge towards when personality traits are ignored. However, we postulate that this is not the ideal personal financial advice to give to all individuals; some customers may be more averse to risk and will thus prefer to avoid volatility in their portfolio. Our personalized agent takes into account such preferences and, e.g., it recommends property investments rather than stock investments.

\begin{figure*}[!ht]
    \centering
    \subfigure[]{
    \includegraphics[width=.48\linewidth]{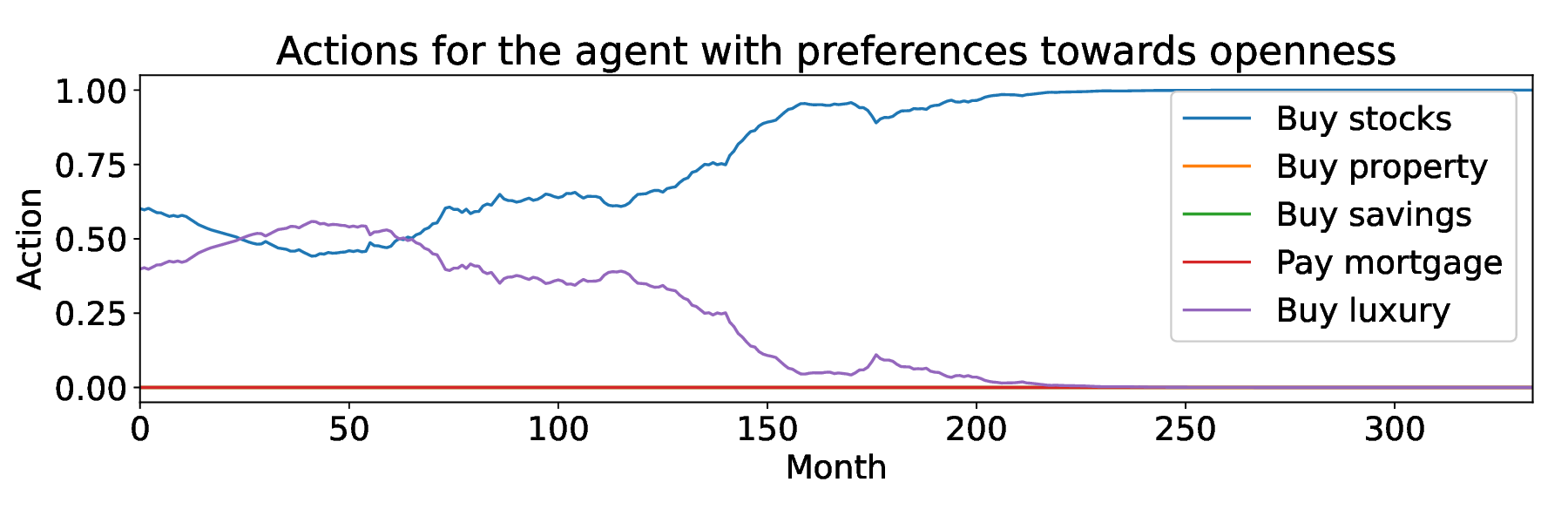}
    }%
    \subfigure[]{
    \includegraphics[width=.48\linewidth]{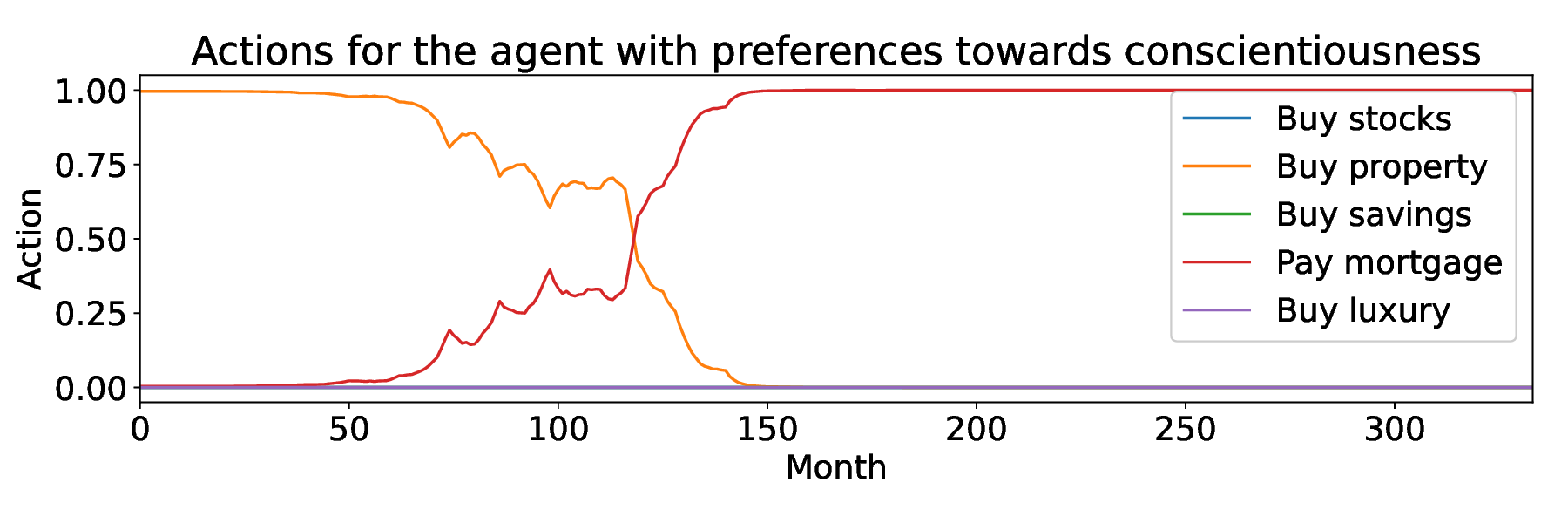}
    }
    \subfigure[]{
    \includegraphics[width=.48\linewidth]{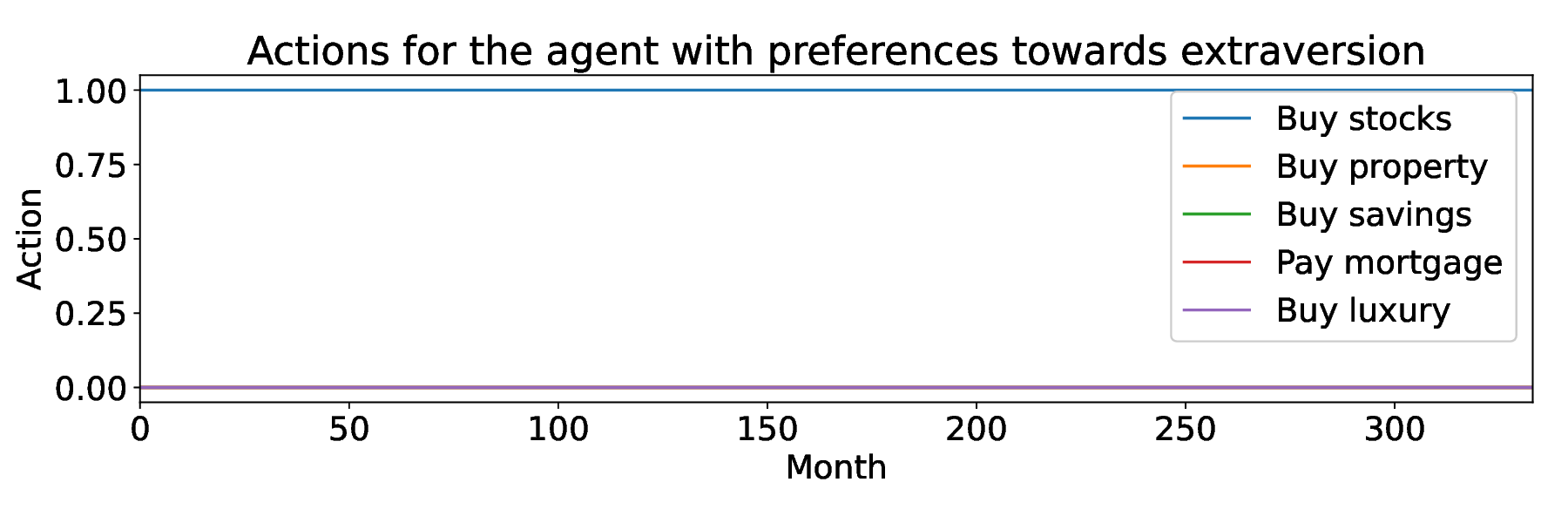}
    }%
    \subfigure[]{
    \includegraphics[width=.48\linewidth]{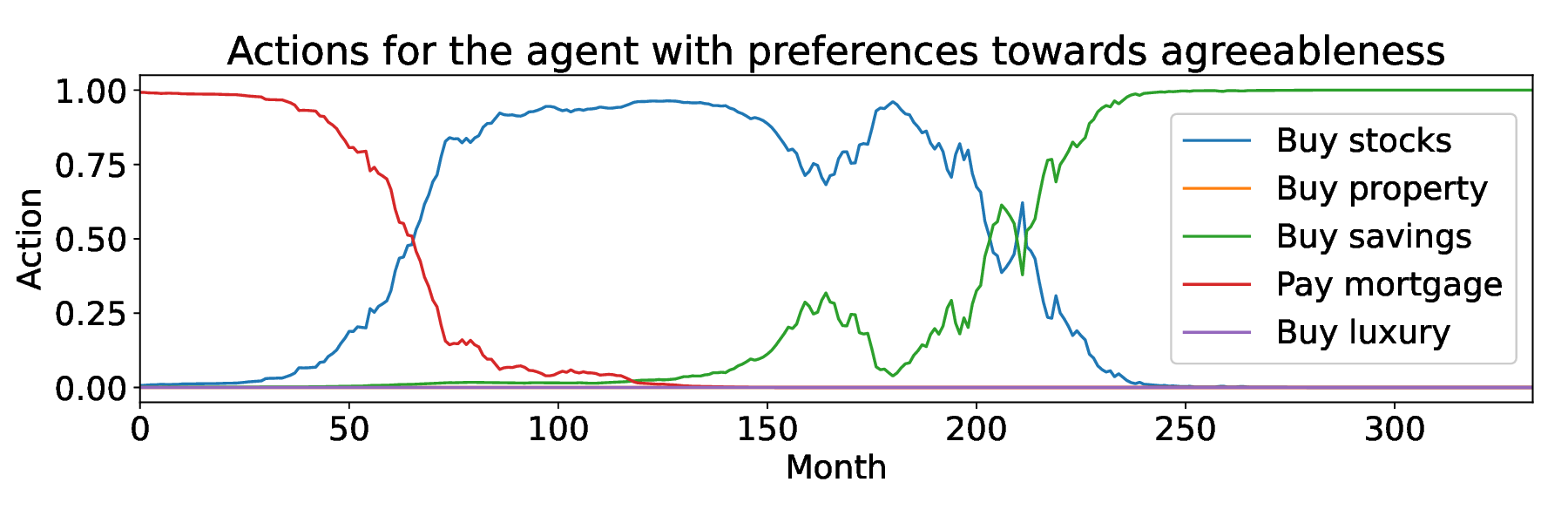}
    }
    \subfigure[]{
    \includegraphics[width=.48\linewidth]{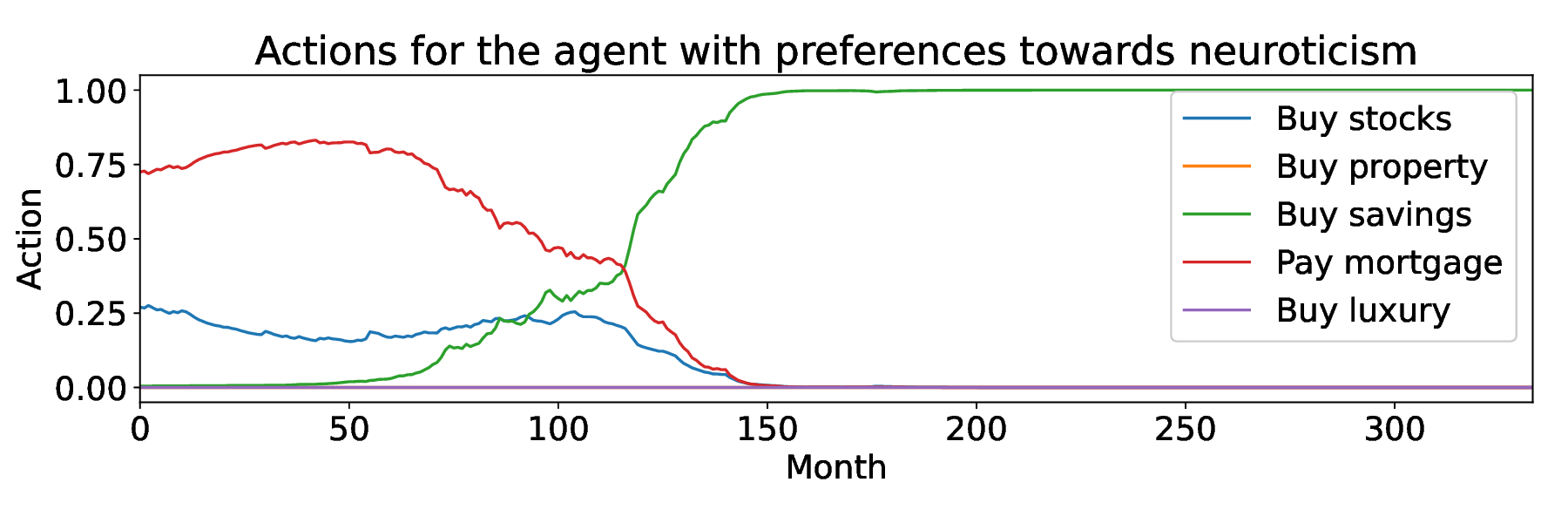}
    }%
    \subfigure[]{
    \includegraphics[width=.48\linewidth]{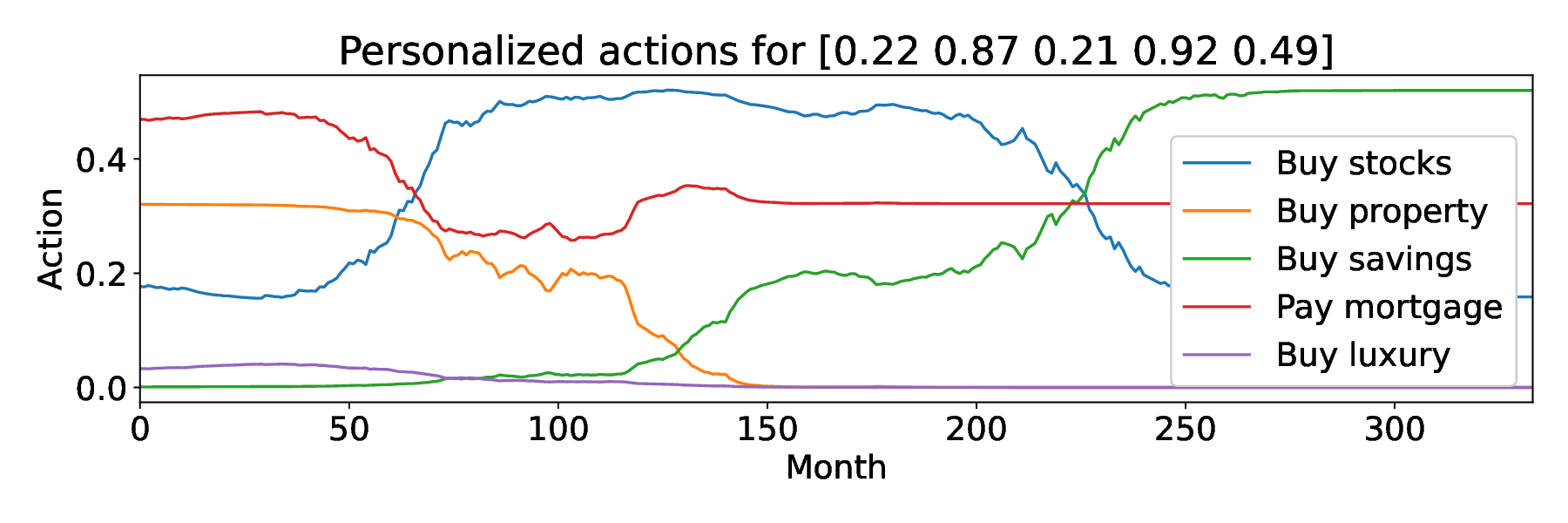}
    }
    \caption{Investment strategies for different prototypical personality traits: Figures (a) through (e) show the fractions of monthly investments for different assets. They reveal the distinct investment strategies with changing asset preferences for the five prototypical personality traits. In Figure (f) we illustrate the investment strategy for a fictitious customer with a random personality profile [openness, conscientiousness, agreeableness, extraversion, and neuroticism] = [0.22, 0.87, 0.21, 0.92, 0.49]. The customer invests in a mixture of assets throughout the investment period.
    }
    \label{fig:agent_actions}
\end{figure*}

Thus far, our agents have each separately learned an optimal investment strategy for each prototypical personality trait. The aggregate policy is the weighted sum of these individually learned policies: a customer has a blend of personality traits which can be represented as a vector with five entries with values within the range $[-1,+1]$. We calculate the inner product of the normalized personality vector and the prototypical policies to arrive at the aggregate investment policy. We show a representative aggregate investment policy for a customer with a random personality profile in Figure~\ref{fig:agent_actions}. We observe that the openness agent is the only agent to recommend spending on luxury items; this is to be expected because its regularisation prior $\pi_0^O$ is the only one with a non-zero coefficient for luxury purchases. We also observe that the conscientiousness agent recommends investing in property in early stages, followed by rigorous loan repayments in the second half of the investment period. This suggests that our agent has learned the concept of compound growth and its utility for portfolio optimization. By contrast, the extraversion agent was steadfast in purchasing stocks only, which is consistent with its regularisation prior $\pi_0^E$. Unlike the conscientiousness agent, the agreeableness and neuroticism agents consistently recommend investing in savings towards the end of the investment period. In the early stages of the investment period, the agreeableness and neuroticism agents utilize compound growth to increase the portfolio value; in the latter phases, their regimen changes and they prefer the safety of savings accounts. This is noteworthy because although risk is not explicitly part of either the reward or regularisation functions, it is consistent with traditional financial advice, which decreases the risk level with age. Repeated training produces consistent results. We intend to elucidate this observation in future work.

\begin{figure*}[!ht]
    \centering
    \subfigure[Learning curve for the regularisation term of the extraversion agent.]{ \label{fig_regterm_1}
    \includegraphics[width=0.4\linewidth]{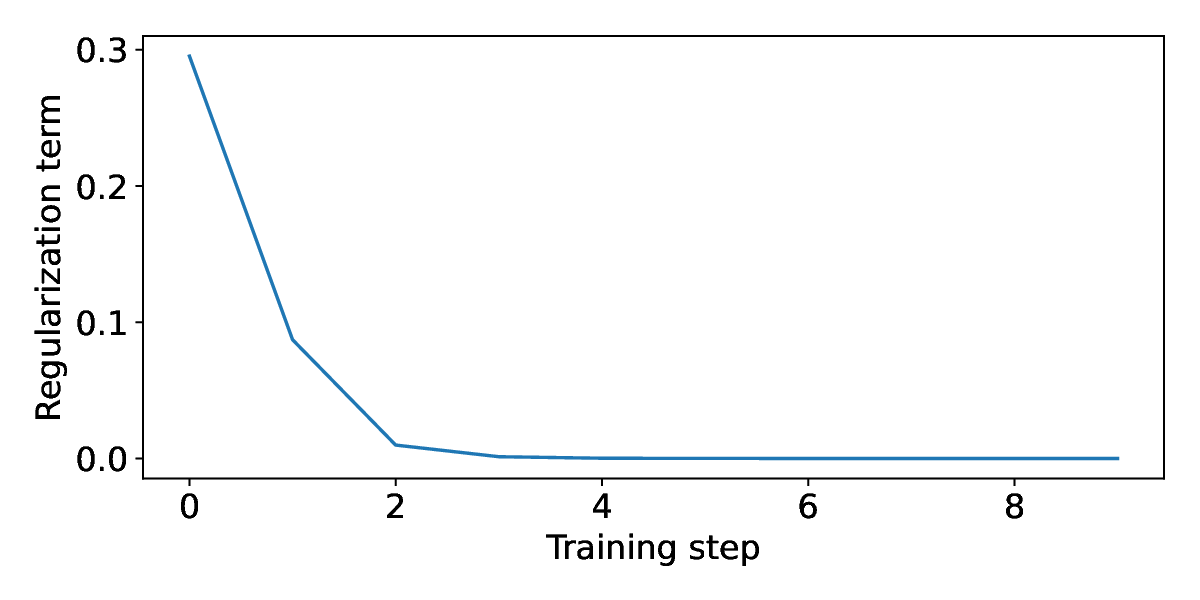}
    }\hspace{0.5cm}
    \subfigure[A typical learning curve for the regularisation term of other agents.]{
    \includegraphics[width=0.4\linewidth]{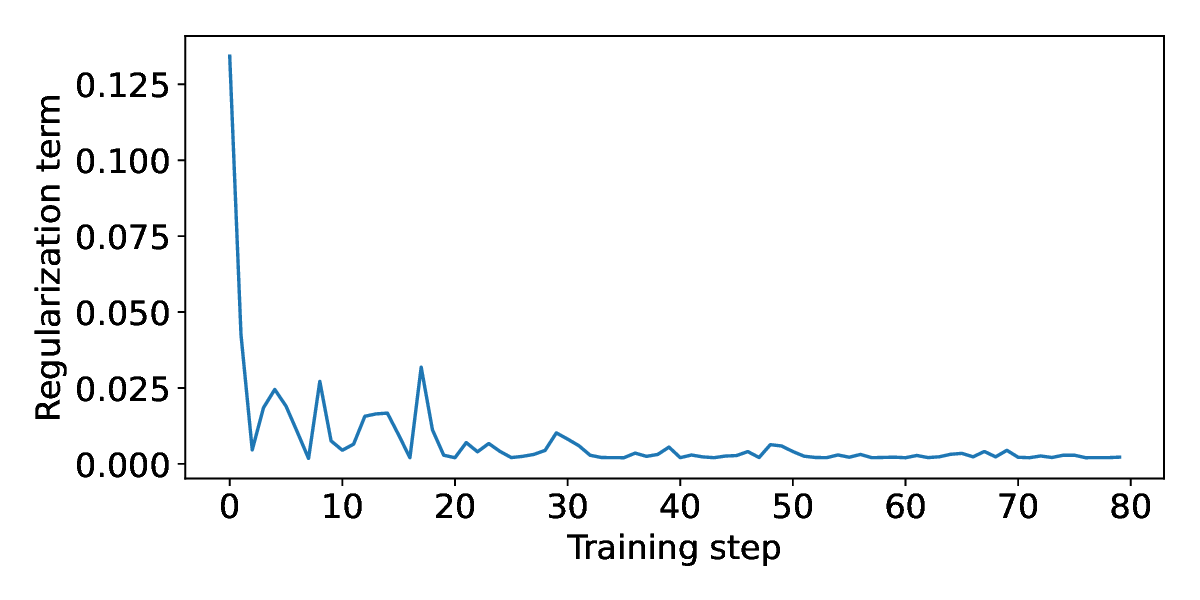}
    }\\
    \subfigure[Learning curve where the regularisation term temporarily falls in a local minimum but subsequently converges to zero.]{
    \includegraphics[width=0.4\linewidth]{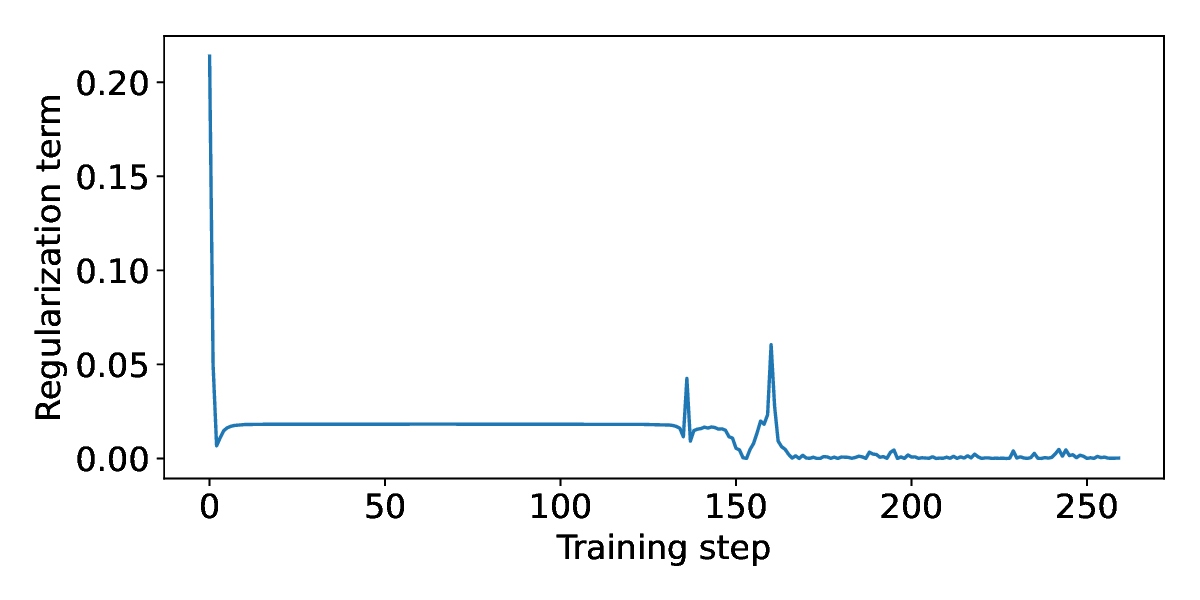}
    }
    \caption{The regularisation term $L$ for three different runs. In (a) the regularisation prior $\pi_0^E$ of the extraverted agent coincides with the optimum monetary policy $\pi^M$ and the policy converges within 5 time steps. (b) shows a typical training run for the other agents which converges within 100-200 training steps. (c) shows a training run where the regularisation term appears to fall in local minimum for a time, but eventually finds the optimum after about 200 training steps. }
    \label{fig:regterm}
\end{figure*}

We observe that training converges quickly to the desired behaviour (see Figure~\ref{fig:regterm}); the contribution of the regularisation term decreases rapidly, which implies that the agent is learning the intended behaviour. We show the regularisation term for the extraversion agent where the regularisation prior $\pi_0^E$ matches the optimum monetary policy $\pi^M$ in Figure \ref{fig_regterm_1}. Further training causes no instability as is often observed in the DDPG algorithm \cite{haarnoja2017}. We hypothesize that this may be due to the agent characteristics imposed by our regularisation whose effect may be similar to entropy regularisation \cite{haarnoja2017}. 

The actions of any linear combination of these agents, i.e., any personal agent, are \emph{interpretable} through the intrinsic characterizations, i.e., priors, of each of the regularized agents. 

%% file: source/conclusions.tex
\section{Conclusions and Directions for Future Work}
We have presented a novel application of training RL agents to exhibit desired characteristics and behaviours in asset management. The method is based on the regularisation of the policy \emph{during} training. Here, we use prototypical personality traits - openness, conscientiousness, agreeableness, extraversion, and neuroticism - to define a set of priors which express their affinity towards different assets and thus impose different investment strategies. This makes the agents' behaviour explicit and thus offers an explanation for their recommendations. Our agents learn distinct optimal strategies for the continuous distribution of monthly investments across a portfolio of investment assets. We have shown that the agents learned to optimize total rewards while adhering to their distinct priors. This makes it possible to interpret the agents' investment strategies.

Unlike traditional DDPG algorithms which may diverge with continuous training, our regularisation results in quick and robust convergence. This could become relevant if RL agents undergo continuous training to give personalized investment advice to customers. The justification of this observation will be subject to future research.

Our agents have learned the concept and utility of compound growth rates and risk avoidance, which form part of the interpretation of their investment strategies. These are solely based on the regularisation priors which express their personality traits; the reward function makes no reference to the personality traits. While the notion of compound growth may emerge from the reward function, we do not yet know whether the notion of risk avoidance is connected to the reward function or regularisation.

Here, we have chosen a linear combination of different, separately trained agents aligned with the prototypical personality traits to arrive at an aggregate investment advice. In the future, we will investigate whether the orchestration of these agents can be learned to approach the optimum monetary policy. This aggregation will need an explanation as well as interpretation to understand its impact on the investment strategy. The hierarchical orchestration of prototypical agents will be learned from real customers' personality profiles. This will result in an explainable and interpretable personalized financial investment advisor.

%% file: source/statements.tex
\vspace{0.7cm}

\subsubsection*{Funding}
This study was partially funded by The Norwegian Research Council; project number 311465.
